\documentclass[conference]{IEEEtran}
\IEEEoverridecommandlockouts
\usepackage{cite}
\usepackage{amsmath,amssymb,amsfonts}
\usepackage{algorithmic}
\usepackage{graphicx}
\usepackage{textcomp}
\usepackage{xcolor}
\usepackage{multirow}
\usepackage{float}
\usepackage{comment}
\usepackage{hyperref}
\usepackage{tabularx,booktabs}
\usepackage{amsfonts}
\usepackage[section]{placeins}
\usepackage[ruled,vlined,linesnumbered]{algorithm2e}
\usepackage{caption}
\usepackage{eso-pic}
\usepackage{booktabs}

\usepackage{mathrsfs}

\usepackage{subfig}

\def\BibTeX{{\rm B\kern-.05em{\sc i\kern-.025em b}\kern-.08em
    T\kern-.1667em\lower.7ex\hbox{E}\kern-.125emX}}
\begin{document}

\title{Benchmarking Deep Learning Classifiers for\\SAR Automatic Target Recognition}

\author{
\IEEEauthorblockN{Jacob Fein-Ashley\IEEEauthorrefmark{1}, Tian Ye\IEEEauthorrefmark{1}, Rajgopal Kannan\IEEEauthorrefmark{2}, Viktor Prasanna\IEEEauthorrefmark{1}, Carl Busart\IEEEauthorrefmark{2}}
\IEEEauthorblockA{
    \IEEEauthorrefmark{1}University of Southern California \IEEEauthorrefmark{2}DEVCOM Army Research Lab\\
    \IEEEauthorrefmark{1}\{feinashl, tye69227, prasanna\}@usc.edu \IEEEauthorrefmark{2}\{rajgopal.kannan.civ, carl.e.busart.civ\}@army.mil}
}

\maketitle

\begin{abstract}
Synthetic Aperture Radar (SAR) Automatic Target Recognition (ATR) is a key technique of remote-sensing image recognition, which can be supported by deep neural networks.
The existing works of SAR ATR mostly focus on improving the accuracy of the target recognition while ignoring the system's performance in terms of speed and storage, which is critical to real-world applications of SAR ATR. For decision-makers aiming to identify a proper deep learning model to deploy in a SAR ATR system, it is important to understand the performance of different candidate deep learning models and determine the best model accordingly.
This paper comprehensively benchmarks several advanced deep learning models for SAR ATR with multiple distinct SAR imagery datasets. Specifically, we train and test five SAR image classifiers based on Residual Neural Networks (ResNet18, ResNet34, ResNet50), Graph Neural Network (GNN), and Vision Transformer for Small-Sized Datasets (SS-ViT). We select three datasets (MSTAR, GBSAR, and SynthWakeSAR) that offer heterogeneity. We evaluate and compare the five classifiers concerning their classification accuracy, runtime performance in terms of inference throughput, and analytical performance in terms of number of parameters, number of layers, model size and number of operations. 
Experimental results show that the GNN classifier outperforms with respect to throughput and latency. However, it is also shown that no clear model winner emerges from all of our chosen metrics and a ``one model rules all" case is doubtful in the domain of SAR ATR.
\end{abstract}


\section{Introduction}
Synthetic Aperture Radar (SAR) Automatic Target Recognition (ATR) is a key technique in remote sensing to recognize on-ground targets with a wide range of applications in real-world scenarios. A SAR system sends radar signals to the ground from the air, processes the reflections, generates high-resolution grayscale SAR images, and passes them to ATR systems for real-time object recognition. 

SAR ATR systems must provide efficient and accurate SAR image classification in real-world applications such as military, agriculture~\cite{sar-agri1,sar-agri2}, and geology. Deep learning techniques have been exploited in ATR systems to classify SAR images. Convolutional Neural Networks (CNNs), which have shown great success in optical image classification~\cite{optical-classification}, are also widely used as SAR image classifiers in ATR systems. They benefit from their strong capability in extracting features of SAR images. Recently, a Graph Neural Network (GNN) model was proposed in our previous paper~\cite{gnn-sar} for SAR image classification. An input SAR image is converted to a grid graph where every pixel as a vertex is connected with its eight neighbors; the SAR image classification is interpreted as a graph classification task. The GNN model was trained on the Moving and Stationary Target Recognition (MSTAR~\cite{mstar,mstar2}) dataset and achieved a state-of-the-art accuracy of 99.1\%.

Although existing CNNs and GNNs have shown high accuracy in the public MSTAR dataset, SAR ATR still faces challenges in their performance in various aspects, such as latency, throughput, and memory storage. The difference in performance will affect the decision to deploy SAR ATR in real-world scenarios. For example, in one case, when fast target recognition is critical, SAR ATR should have low latency and high throughput. In contrast, in another case, the model size is more essential when memory is a limited resource in the system. Therefore, it is necessary to perform a comprehensive benchmarking to compare the performance of different deep learning models for SAR ATR so that a proper model can be chosen to deploy according to the requirements and constraints of the application. 

In this paper, we benchmark deep learning-based classifiers for SAR ATR. We evaluate the performance of SAR image classifiers in terms of their classification accuracy, runtime performance, and analytical performance. The runtime version includes inference latency and throughput that represents the speed of the ATR system. The analytical performance has the size of the model, the number of parameters, the number of layers, and the number of operations, representing resource consumption and efficiency. In this empirical study, we evaluate and compare three classes of typical deep learning models: Residual Neural Network (ResNet)~\cite{DBLP:journals/corr/HeZRS15}, Graph Neural Network (GNN)~\cite{gnn-sar}, and a variant of the Vision Transformer (ViT)~\cite{dosovitskiy2021an}. The ViT (SS-ViT) variant allows for higher accuracy on a smaller dataset, proposed in~\cite{smallvit}. 
To cover more real-world scenarios, we select three distinct datasets with significant diversity as summarized in Table~\ref{tab:dataset}. 
\begin{table}[!htbp]
    \caption{Selected SAR Imagery Datasets}
    \label{tab:dataset}
    \centering 
    \resizebox{\linewidth}{!}{
        \begin{tabularx}{1.1\linewidth}{@{} c| *{3}{c} l @{}}
        \toprule
         & \textbf{MSTAR~\cite{mstar2}} & \textbf{GBSAR~\cite{Turčinović_2023}} & \textbf{SynthWakeSAR~\cite{SynthWakeSAR}}  \\
        \midrule
        SAR device & Airborne-based & Ground-based & Airborne-based \\
        Target object & Military vehicles & Ceramic objects & Ship wakes \\
        Data type & Images & Raw radar data & Images\\
        
        Training size & $27,000$ & $5,147$ & $36,864$  \\
        Testing size & $2,425$ & $1,287$ & $9,216$ \\
        \# classes & $10$ & $7$ & $10$ \\        
        \bottomrule
        \end{tabularx}
    }
\end{table}
The three datasets are
Moving and Stationary Target Recognition (MSTAR)~\cite{mstar2}, Ground-Based SAR Data Obtained With Different Combinations of Bandwidth and Step Size (GBSAR)~\cite{Turčinović_2023}, and a dataset containing various simulated ship wakes (SynthWakeSAR)~\cite{SynthWakeSAR}. Section~\ref{sec:benchmark} describes more details about the datasets.


Our main contributions are summarized as follows:
\begin{itemize}
    \item We conduct a benchmarking study that compares the classification accuracy, runtime performance, and analytical performance of SAR ATR based on three types of deep learning models (ResNet, GNN, and SS-ViT) with three distinct SAR imagery datasets (MSTAR, GBSAR, and SynthWakeSAR).
    \item Results tend to be mixed with respect to our chosen metrics. We observe that the GNN we proposed earlier~\cite{gnn-sar} wins in terms of the number of multiply/accumulate operations, the number of model layers (these metrics are explained in Section~\ref{sec:benchmark}), and accuracy for the MSTAR dataset. The SS-ViT outperforms in the number of model parameters and size but underperforms in accuracy. We notice that the ResNet architectures contain an enormous parameter and the number of MAC operations but are more accurate than the other models in the GBSAR and SynthWakeSAR datasets.
    \item This work finds that, like many machine learning problems, there is still ``no free lunch" 
    when it comes to a single model outperforming others in terms of all performance metrics.
    Additionally, we find that a more thorough assessment of deep learning models for SAR ATR requires evaluating the models across a variety of datasets. For this purpose, it is imperative to develop more public SAR datasets.

\end{itemize}

\section{Background and Related Work}
\subsection{SAR and SAR ATR}

Synthetic Aperture Radar (SAR) imagery is a radar imaging technique with wide applications in military~\cite{sar-military}, climate studies~\cite{sar-climate}, marine surveillance
~\cite{sar-marine}, and agriculture~\cite{sar-agri1,sar-agri2}. A SAR system, typically equipped on satellites or aircraft, transmits radar signals toward the earth's surface. After receiving reflections from the earth, the system generates a grayscale image depending on the backscatter intensity of the reflections. Compared to optical sensors, radar signals are independent of lighting and weather conditions, such as clouds. As a result, SAR imagery can obtain high-resolution images of the earth in all weather states and at all times of day and night. 

SAR Automatic Target Recognition (ATR) automatically identifies the objects in a SAR image, which can be interpreted as a task of image classification. Different from classifying optical images, SAR image classification has several challenges: (1) SAR images are single-channel grayscale images representing the intensity of backscattered radar signal, which carries less information than RGB images with three channels of colors. (2) SAR images of the same object can appear differently when they are imaged 
with different \textbf{capturing parameters} such as radar frequency, azimuth, incidence angle, and resolutions. For example, incidence angles will affect the appearance of shadow regions in SAR images and make the brightness of a surface different.
(3) Large-scale public labeled datasets of SAR images are hard to access~\cite{lacksardatasets} and therefore it is difficult to develop effective classifiers for unseen SAR images.
Despite these challenges, many classifiers based on deep learning models that have been proven successful in optical image classification can still be applied to the domain of SAR ATR.


\subsection{Deep Learning Models for SAR ATR}
\label{sec:models}

Convolutional Neural Networks (CNNs) are a class of most typical models for image classifications. They apply a series of convolutional kernels to local regions of input images, which leads to a strong ability to capture local features. Residual Networks (ResNets) are a type of CNNs that are usually easier to train and show higher accuracy than standard CNNs. There are several versions of ResNets with different numbers of layers. We consider three versions with fewest layers (ResNet18, ResNet34 and ResNet50) in this paper because more layers will not further improve the accuracy in our experiments.

Unlike CNNs, SAR ATR based on Graph Neural Networks (GNNs) interprets target recognition as a graph classification task. In our earlier work~\cite{gnn-sar}, we proposed a SAR ATR based on GNN. A given SAR image can be converted into a graph: Each pixel becomes a vertex of the graph with the pixel value as the vertex attribute, and a non-attributed edge connects every pair of neighbor pixels. 
\begin{figure}[!htbp]
\centering
\includegraphics[width=0.3\textwidth]{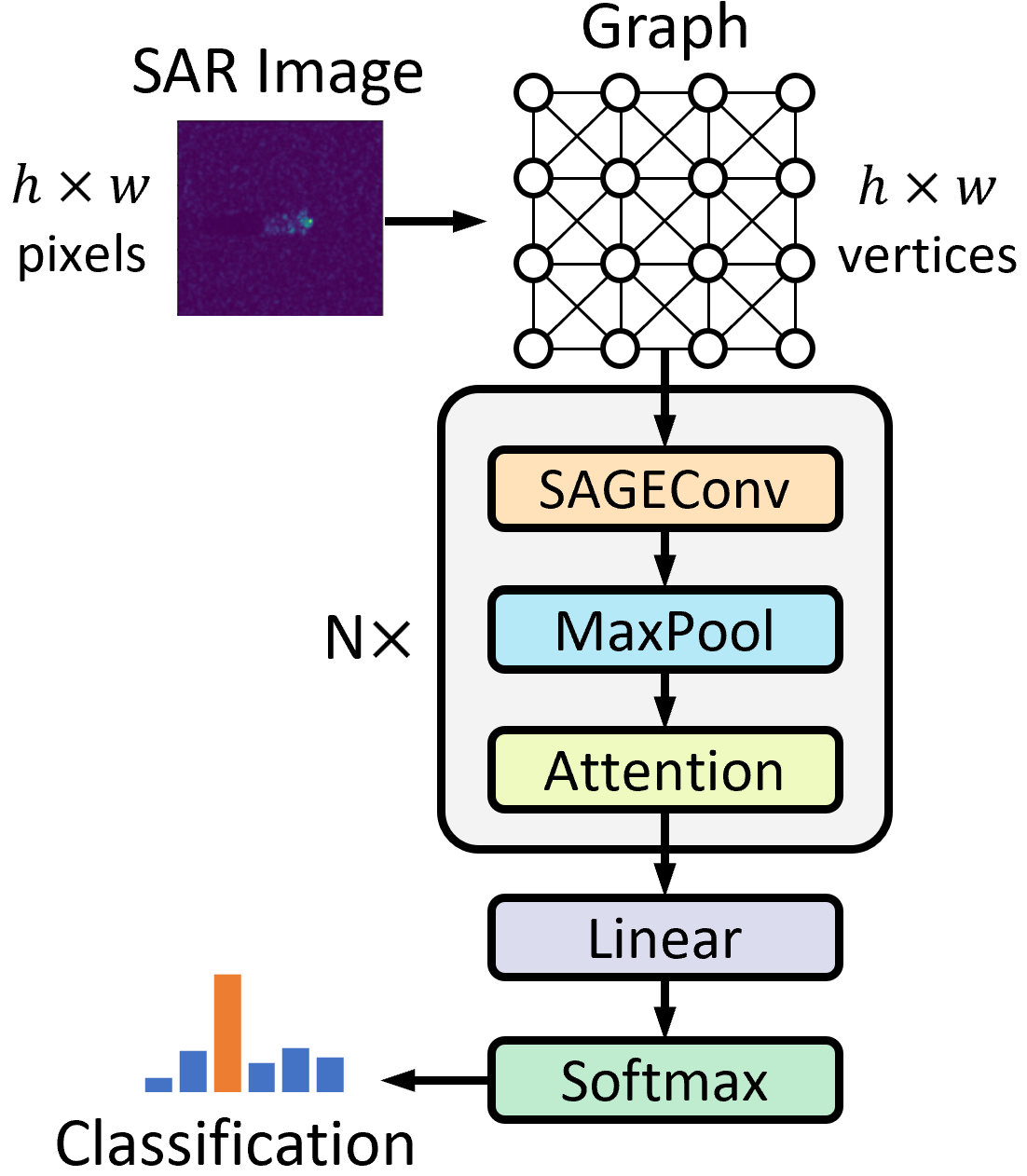}
\caption{Architecture of GNN-based SAR image classifier}
\label{fig:gnn-arch}
\end{figure}
Figure~\ref{fig:gnn-arch} shows the architecture of the GNN-based SAR image classifier~\cite{gnn-sar}. $N$ sequential groups of feature extractors generate an embedding for the graph, followed by a linear layer (``Linear") and softmax layer (``Softmax") that outputs the classification result. Each group of feature extractors consists of a GraphSAGE convolution operator~\cite{graphsage} (``SAGEConv") that generates embeddings for each vertex by aggregating the features of neighbors vertices, a max pooling layer (``MaxPool") to reduce the dimension of the embeddings, and a layer of attention mechanism~\cite{attention} (``Attention") to guide the classifier to focus on the most important parts of the embeddings. In the experiments of our previous paper~\cite{gnn-sar}, the GNN classifier obtains an accuracy of over 99\% on the MSTAR dataset. Compared with CNNs, GNNs have lower computational costs while still achieving a similar level of classification accuracy. 

Vision Transformer (SS-ViT~\cite{dosovitskiy2021an,smallvit}) is a recently proposed technique that takes a different approach from CNNs and GNNs. The previous two classes of models interpret the input image as two-dimensional pixels: CNNs treat the input image as a matrix of pixel values, and GNNs build the graph by mapping each pixel as a vertex. Instead, SS-ViT divides an image into a sequence of small patches and then feeds the patches into a transformer model. The transformer can understand the relative importance of the patches and the relationship between patches. Therefore, SS-ViT has a better global understanding of the input image. Compared to CNNs and GNNs, the limitations of SS-ViT include its high requirements on the amount of training data and computational resources. To remedy this, \cite{smallvit} proposed a shift patch tokenization (SPT) and locality self-attention (LSA) alteration to the originally proposed SS-ViT. The change in the SS-ViT structure will allow us to use recent transformer machine-learning models on smaller datasets.

\section{Benchmarking Deep Learning Models for SAR ATR}
\label{sec:benchmark}
Our motivations for a benchmark of deep learning models for heterogeneous SAR ATR datasets are two-fold:
(1) With the recent release of SAR ATR datasets for classification purposes, we seek to study whether different datasets will influence the performance of deep learning models. 
(2) As many deep learning models are developed, we intend to compare them concerning multiple metrics, which will help practitioners to determine what model to use in a SAR ATR system.

In the space of SAR ATR deep learning, the MSTAR dataset is widely used. 
However, only relying on a single dataset is insufficient to assess the performance of the deep learning models on unseen SAR data.
We hypothesize that deep learning models will depend on a mixture of SAR capturing methods, parameters, azimuth angles, and elevation criteria. We source from the limited supply of publicly available SAR datasets. From these datasets, we show the performance of a recently constructed GNN, ResNet18, ResNet34, ResNet50, and SS-ViT. The SS-ViT and ResNet models are widely available and often used for high-performing image classification tasks. Benchmarking performance in this variety of models for heterogeneous datasets is a vast frontier, as the number of publicly available datasets is limited. Exploring this frontier may expose the importance and sensitivity of SAR capturing parameters in the deep learning space.

\subsection{Datasets} 

We obtain a variety of datasets to provide a varied perspective on azimuth, elevation, and range-to-velocity ratio, which are often factors in SAR image degradation.  To retain consistency, all processed datasets require a resizing method to ($88 \times 88$) pixels~\cite{preprocessing}. SAR ATR datasets are seldom released to the public due to data sensitivity and acquisition difficulty. Thus, measuring popular deep learning models on SAR datasets with differing properties respective to azimuth, elevation, and SAR measuring device parameters is critical~\cite{Zhao_Huang_Xin_Guo_Pan_2021}. Details of the selected datasets and their sample images are as follows:
\begin{enumerate}
    \item[1.] \textbf{MSTAR}~\cite{mstar2} is a widely known SAR dataset that is publicly available and captures images at varying elevations and azimuth angles with $10$ classes.
    \begin{figure}[H]
        \centering
\includegraphics[width=0.45\textwidth]{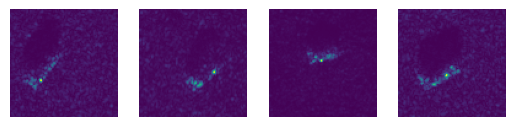}
        \caption{Sample MSTAR Dataset Images}
        \label{fig:mstar}
    \end{figure}
    Concerning capturing parameters, all are constant except for elevations, which range from $14^{\circ} - 18^{\circ}$~\cite{Lewis_Levy_Nehrbass_Scarnati_Sudkamp_Zelnio}.
    
    \item[2.] \textbf{SynthWakeSAR}~\cite{SynthWakeSAR} contains synthetic (simulated) SAR images of wakes from $10$ visible ships, consisting of normalized radar cross-sections (NRCS) with tilt and hydrodynamic modulations and velocity bunching. SynthWakeSAR releases distributions of noise-free images, noisy images with simulated speckles, and despeckled SAR images. For our purposes, we train our models on denoised images, as recommended by the authors of SynthWakeSAR~\cite{SynthWakeSAR}. The authors of this dataset do not state their capturing parameters for this dataset but mention that each scene remains the same size.

    \begin{figure}[H]
        \centering
        \includegraphics[width=0.45\textwidth]{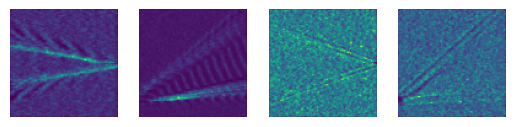}
        \caption{Sample SynthWakeSAR Dataset Images}
        \label{fig:synthwakesar}
    \end{figure}
    
    \item[3.] \textbf{GBSAR}~\cite{Turčinović_2023} captures ceramic cups with rubber objects, each captured with different step sizes and bandwidths. GBSAR contains $7$ classes. Note that GBSAR is a dataset of raw SAR data instead of reconstructed images. As suggested by the authors of GBSAR, we implement end-to-end learning by directly feeding the raw SAR data into the models. The capturing parameters of GBSAR vary, with step sizes $\in \{0.5, 1, 1.5, 3, 3.5\}$ cm, and three bandwidth cases $\in \{300, 600, 1300\}$ MHz.

    \begin{figure}[H]
        \centering
        \includegraphics[width=0.45\textwidth]{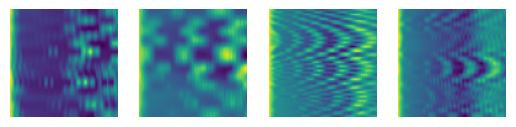}
        \caption{Sample SynthWakeSAR Dataset Images}
        \label{fig:gbsar}
    \end{figure}
\end{enumerate}

\subsection{Metrics}
In addition to selecting datasets containing mixed capturing parameters, it is important to choose metrics that fairly compare results concerning runtime, accuracy, and practicality. The particular metrics we consider are as follows:
\begin{enumerate}
    \item[1.]
\textbf{Throughput and latency} allow for comparing the inference time  across various models. Throughput is the maximum number of input instances a model can process per time unit. For our purposes, we define throughput as
\begin{align*}
    \text{Throughput} &= \frac{\text{Total number of images processed}}{\text{Total inference time}}
\end{align*}
Thus, it is natural that the throughput units are in images processed per time unit. 
We also define \textbf{latency} as the average time of completion for a classification to be completed by a model.

\item[2.] Additionally, we use the \textbf{accuracy} metric to assess how well each classifier performs during prediction. 

\item[3.] Other metrics we consider are analytical performance, including the \textbf{number of parameters (\# Parameters)}, \textbf{model size}, \textbf{layer count (\# Layers)}, and \textbf{number of multiply/accumulate operations (\# MACs)}. In other words, suppose accumulator $a$ counts an operation of arbitrary $b,c \in \mathbb{R}$.  We count the number of operations as $a \leftarrow a + (b \times c)$. The layer count metric is an important factor of latency. Increasing the number of layers will also increase the latency of an inference of a model. The goal of an effective machine learning model is to maximize throughput while minimizing the number of MACs and a number of layers, in our case.  
\end{enumerate}
In summary, accuracy represents the ability of a model to recognize objects.
Throughput, model size, \# Parameters, and \# MACs allow us to assess the practicality of each model. \# Layers are one factor that can affect the inference latency and the potential capacity for hardware acceleration.

\subsection{Deep Learning Models} A direct comparison of state-of-the-art models is conducted by comparing our GNN architecture to ResNet18, ResNet34, ResNet50, and SS-ViT. Baseline models are selected by heuristic performance on popular image classification datasets. ResNet architectures and SS-ViT are state-of-the-art models with well-known high accuracy for standard image classification tasks. Details of the models are described in Section~\ref{sec:models}.

\section{Experiments}
\label{section:experiments}

\subsection{Experimental Setup}

We implement the training and testing of the three types of deep learning models using the PyTorch 1.11.0~\cite{pytorch} library on a state-of-the-art server with CUDA 11.8 powered by an NVIDIA RTX A6000 GPU. The implementation of ResNet18, ResNet34, ResNet50 and SS-ViT are based on open-source repositories~\cite{torchvision2016,vit-github}
. We use the PyTorch-OpCounter~\cite{thop} library to count \# Parameters and \# MACs. The datasets, metrics, and deep learning models are summarized in the previous section.

\subsection{Performance Comparison}

We first compare the analytical performance using \# MACs, \# Parameters, model size, and \# Layers for each model. As shown in Table~\ref{macsparam}, among the studied deep learning models, GNN has the fewest layers and fewest operators in one forward propagation. Therefore, it is easier for potential hardware accelerators such as FPGAs and ASICs to improve the performance. The SS-ViT has the fewest parameters and smallest model size that better fits systems when memory is a critical resource.

\begin{table}[H]
    \caption{Analytical Performance}
    \label{macsparam}
    \centering 
        \begin{tabularx}{\linewidth}{@{} c *{5}{c} c @{}}
            \toprule
            \textbf{Model} & \textbf{$\#$ MACs} & \textbf{$\#$ Parameters} & \textbf{Model Size (Mb)} & \textbf{$\#$ Layers}\\
            \midrule
            ResNet18 & $1.87 \times 10^{10}$ & $1.12 \times 10^7$ & $42.67$ & $53$\\
            ResNet34 & $3.92 \times 10^{10}$ & $2.13 \times 10^7$ & $81.26$ & $93$ \\
            ResNet50  & $4.41 \times 10^{10}$ & $2.35 \times 10^7$ & $89.93$ & $127$\\
            SS-ViT  & $1.70 \times 10^{10}$ & $\bf{5.15 \times 10^5}$ & $\bf{2.12}$ & $247$ \\
            GNN  & $\bf{8.15 \times 10^7}$ & $1.27 \times 10^6$ & $5.08$ & $\bf{36}$\\

            \bottomrule
        \end{tabularx}
\end{table}

Additionally, we measure the various metrics for classification accuracy and throughput for each dataset mentioned as shown in Tables ~\ref{mstartab}, ~\ref{synthtab}, ~\ref{gbsartab}. In most cases, our GNN outperforms in terms of  throughput and latency with the exception that ResNet18 achieves better throughput on the SynthWakeSAR dataset. Each dataset yields a distinct deep learning model as the most accurate. No single model achieves better accuracy than others across all the three datasets.

\begin{table}[H]
    \caption{MSTAR Performance}
    \label{mstartab}
    \centering 
        \begin{tabularx}{0.875\linewidth}{@{} c *{3}{c} l @{}}
        \toprule
        \textbf{Model} & \textbf{Accuracy} & \textbf{Throughput ($\text{imgs}/\text{ms}$)} & \textbf{Latency (ms)}\\
        \midrule
        ResNet18 & $98.47\%$ & $5.91  \times 10^2$ & $1.87 \times 10^{-3}$ \\
        ResNet34 & $98.64\%$ &$4.96  \times 10^2$  & $1.94 \times 10^{-3}$ \\
        ResNet50 & $90.34\%$ &$4.50 \times 10^2$ & $2.12 \times 10^{-3}$  \\
        SS-ViT & $95.61\%$ & $4.84  \times 10^2$ & $2.07 \times 10^{-3}$ \\
        GNN & $\bf{99.09\%}$ & $\bf{8.56 \times 10^2}$ & $\bf{1.16 \times 10^{-3}}$   \\
        \bottomrule
        \end{tabularx}
\end{table}

\begin{table}[H]
    \caption{Denoised SynthWakeSAR Performance}
    \label{synthtab}
    \centering 
        \begin{tabularx}{0.875\linewidth}{@{} c *{3}{c} l @{}}
        \toprule
        \textbf{Model} & \textbf{Accuracy} & \textbf{Throughput ($\text{imgs}/\text{ms}$)} & \textbf{Latency (ms)}\\
        \midrule
        ResNet18 & $90.30\%$ & $\bf{1.82 \times 10^3}$ & $5.56 \times 10^{-4}$ \\
        ResNet34 & $92.14\%$ & $1.64 \times 10^3$ & $6.12 \times 10^{-4}$\\
        ResNet50 & $\bf{92.42}\%$ & $ 1.54 \times 10^3$ & $7.80 \times 10^{-4}$ \\
        SS-ViT & $87.98\%$ & $0.67 \times 10^3$ &  $14.66 \times 10^{-4}$  \\
        GNN & $91.15\%$ &  $1.75 \times 10^{3}$ &  $\bf{4.55 \times 10^{-4}}$ \\
        \bottomrule
        \end{tabularx}
\end{table}

\begin{table}[H]
    \caption{GBSAR Performance}
    \label{gbsartab}
    \centering 
         \begin{tabularx}{0.875\linewidth}{@{} c *{3}{c} l @{}}
        \toprule
        \textbf{Model} & \textbf{Accuracy} & \textbf{Throughput ($\text{imgs}/\text{ms}$)} & \textbf{Latency (ms)}\\
        \midrule
        ResNet18 & $99.30\%$ & $3.02 \times 10^2$ & $3.29 \times 10^{-3}$ \\
        ResNet34 & $\bf{99.99}\%$ & $2.98 \times 10^2$ & $3.33 \times 10^{-3}$\\
        ResNet50 & $99.53\%$ & $2.84 \times 10^2$ & $3.62 \times 10^{-3}$\\
        SS-ViT & $99.04\%$ & $3.59 \times 10^2$ & $1.84 \times 10^{-3}$  \\
        GNN & $98.67\%$ & $\bf{5.43 \times 10^2} $ & $\bf{1.79 \times 10^{-3}}$\\
        \bottomrule
        \end{tabularx}
\end{table}

\subsection{Discussions}

\textbf{Insight 1:} No single model can outperform across all provided metrics according to our empirical measurement. Regarding accuracy, the ResNet architectures outperform the SS-ViT and GNN with the SynthWakeSAR and GBSAR dataset cases. Our GNN classifier transcends the other models with respect to throughput and latency across all datasets. Considering the results of the inference time, model complexity, size, and performance in the machine-learning application is essential. Thus, large learning models like ResNet may not be practical. The fastest model concerning inference time in our set of models and datasets is the GNN.

\textbf{Insight 2:} We recommend that the practitioner thoroughly analyze their dataset before deploying deep learning models for SAR ATR usage. Due to the sensitivity of SAR capturing equipment, having a ``single model rule all" case is unrealistic. As empirically evaluated in this paper, it is imperative to consider the implications of deploying an enormous model that delivers only marginally better accuracy. We leave it to the practitioner to perform a cost-benefit analysis when deploying deep learning models for SAR usage.

\textbf{Insight 3:} With the limited public release of SAR imaging data, predicting how models will perform on datasets is difficult. Realistically, providing SAR images with azimuth angles, elevation, and SAR capturing parameters may assist models to become more robust. Heuristically, it is seen that, although simple, the GBSAR dataset contains mixed parameters concerning bandwidth and step size. All models perform quite well on this dataset, so obtaining datasets with varying parameters, such as GBSAR, may be necessary to strengthen future deep learning models.

\textbf{Insight 4:} As simulated datasets become more prevalent, as seen in SynthWakeSAR, future study areas may include synthesizing SAR ATR datasets with various capturing parameters, e.g., azimuth angles and elevations. Synthesizing a dataset with diverse parameters, speckling, rotation, and other mentioned capturing methods would help create models invariant to rotation, speckling, and noisy imaging, a prevalent topic in the SAR ATR realm.

\section{Conclusion}
This work benchmarked the performance of widely popular deep learning image classification models with respect to various accuracy and computational performance metrics. We highlighted the necessity of data evaluation from the practitioner to create a robust model. We found that with the models we had experimented with, the practitioner must make informed decisions when selecting a model. The practitioner must perform a well-informed cost-benefit analysis with computational complexity and accuracy in mind.

The field of SAR ATR is exceedingly dependent on the MSTAR dataset. We believe heterogeneous datasets will be important in advancing the deep learning SAR ATR realm and future SAR image classification models.

We will consider the following future directions to improve further the robustness of deep learning models for SAR ATR. On the one hand, deep learning models should be designed to remain unaffected by variations in SAR capturing parameters. For example, the models should recognize objects in SAR images from various azimuth angles and elevations. Also, as SAR images inevitably contain “salt and pepper” like noise spreading all over the images, deep learning models should be invariant to the speckle. On the other hand, deep learning models should be robust against perturbations from potential adversaries. In our paper~\cite{spie-gnnattack}, we have studied the vulnerability of our GNN model in the face of adversarial attacks. The GNN model's accuracy degrades when the SAR images' pixel values are slightly manipulated. This shows the necessity of more robust deep learning models for SAR ATR. In the future, by studying more advanced adversarial perturbations and developing defensive techniques accordingly, the robustness of deep learning models will be significantly improved. Additionally, this experiment was run using only one state-of-the-art GPU. Further studies may include experimenting with different GPUs and train-to-test ratios. 

\section*{Acknowledgement}

 \textbf{Distribution Statement A:} Approved for public release. Distribution is unlimited.

 This work is supported by the DEVCOM Army Research Lab (ARL) under grant W911NF2220159.

\bibliographystyle{IEEEtran}
\bibliography{SAR-ATR}

\end{document}